\documentclass[conference]{IEEEtran}
\IEEEoverridecommandlockouts
\usepackage{cite}
\usepackage{amsmath,amssymb,amsfonts}
\usepackage{algorithmic}
\usepackage{graphicx}
\usepackage{textcomp}
\usepackage{xcolor}
\def\BibTeX{{\rm B\kern-.05em{\sc i\kern-.025em b}\kern-.08em
    T\kern-.1667em\lower.7ex\hbox{E}\kern-.125emX}}

\usepackage{booktabs}

\usepackage{multirow}

\definecolor{ao(english)}{rgb}{0.0, 0.5, 0.0}

\newcommand{\green}[1]{{\textcolor{ao(english)}{#1}}}

\usepackage{fancyhdr}
\begin{document}

\title{
Investigating Enhancements to Contrastive Predictive Coding for Human Activity Recognition
}

\author{\IEEEauthorblockN{Harish Haresamudram}
\IEEEauthorblockA{\textit{School of Electrical and Computer Engineering} \\
\textit{Georgia Institute of Technology}\\
Atlanta, USA \\
}
\and
\IEEEauthorblockN{Irfan Essa}
\IEEEauthorblockA{\textit{School of Interactive Computing} \\
\textit{Georgia Institute of Technology}\\
Atlanta, USA \\
}
\and
\IEEEauthorblockN{Thomas Pl\"otz}
\IEEEauthorblockA{\textit{School of Interactive Computing} \\
\textit{Georgia Institute of Technology}\\
Atlanta, USA \\
}
}

\maketitle

 \thispagestyle{fancy}
\fancyhead{} 
\fancyfoot{} 
\pagenumbering{gobble}
\fancyfoot[C]{\textcolor{red}{This manuscript is under review. Please write to hharesamudram3@gatech.edu for up-to-date information}}

\begin{abstract}
The dichotomy between the challenging nature of obtaining annotations for activities, and the more straightforward nature of data collection from wearables, has resulted in significant interest in the development of techniques that utilize large quantities of unlabeled data for learning representations. 
Contrastive Predictive Coding (CPC) is one such method, learning effective representations by leveraging properties of time-series data to setup a contrastive future timestep prediction task.
In this work, we propose enhancements to CPC, by systematically investigating the encoder architecture, the aggregator network, and the future timestep prediction, resulting in a fully convolutional architecture, thereby improving parallelizability. 
Across sensor positions and activities, our method shows substantial improvements on four of six target datasets, demonstrating its ability to empower a wide range of application scenarios. 
Further, in the presence of very limited labeled data, our technique significantly outperforms both supervised and self-supervised baselines, positively impacting situations where collecting only a few seconds of labeled data may be possible.
This is promising, as CPC does not require specialized data transformations or reconstructions for learning effective representations.
\end{abstract}


\label{sec:intro}
\section{Introduction}
Sensor-based Human Activity Recognition (HAR), which involves the automatic prediction of activities being performed, has a multitude of applications, including health \cite{bachlin2009wearable, morshed2022personalized, morshed2019prediction}, industry \cite{stiefmeier2008wearable, yoshimura2022acceleration, morales2022acceleration}, and fitness tracking \cite{liaqat2019wearbreathing, banos2014mhealthdroid}.
However, this process is highly dependent on the availability of annotated data, which is typically collected in laboratory settings and for orchestrated sets of activities, resulting in limited size and diversity of activities and participants \cite{kwon2020imutube}.

The ubiquity of sensors such as accelerometers and gyroscopes on commodity devices such as smartwatches and smartphones, allows for the straightforward collection of large quantities of unlabeled movement data. 
For example, one can be asked to wear an Apple Watch for collecting data for a period of time, resulting in data from true ``in-the-wild" conditions. 
However, annotating this data is not as straightforward because it is typically considered privacy invasive, cumbersome, and sometimes simply impossible from a logistical perspective \cite{plotz2018deep}. 
This dichotomy between the challenging nature of annotating activities and the more straightforward nature of data collection has resulted in the introduction of modeling techniques that can leverage large quantities of unlabeled data to first learn representations, that are subsequently fine-tuned to the specific activities of interest.

This training paradigm of `pretrain-then-finetune' has resulted in significant performance improvements across domains such as computer vision \cite{gidaris2018unsupervised}, speech processing \cite{chung2019unsupervised, chung2020generative, chung2020improved, schneider2019wav2vec}, natural language processing \cite{devlin2018bert, pennington2014glove, mikolov2013efficient}, and human activity recognition from wearables \cite{saeed2019multi, haresamudram2020masked, haresamudram2021contrastive}. 
Going beyond traditional unsupervised learning methods such as Restricted Boltzmann Machines (RBM) \cite{plotz2011feature} and Autoencoders \cite{haresamudram2019role}, many methods have been proposed for learning self-supervised representations for wearables, including Multi-task self-supervision \cite{saeed2019multi}, Masked Reconstruction \cite{haresamudram2020masked}, SimCLR \cite{tang2020exploring}, and Contrastive Predictive Coding (CPC) \cite{haresamudram2021contrastive}.
These methods have reported great improvements in performance relative to supervised baselines, specifically on target datasets containing locomotion-style activities.

In this paper, we focus on the Contrastive Predictive Coding framework, which has been utilized for tasks such as representation learning \cite{haresamudram2021contrastive} and change point detection \cite{deldari2021time} on wearable sensor datasets.
The pretext task involves future timestep prediction in a contrastive learning setup, which leverages properties of time-series data (e.g., the ability to predict future timesteps from the past) for effective training. 
Such a framework is attractive as it does not require the design of specialized data transformations (like in the case of SimCLR \cite{chen2020simple} for example), which can be dependent on the data streams utilized, or reconstruction of (parts of) the input data (e.g., Autoencoders \cite{haresamudram2019role} and Masked reconstruction \cite{haresamudram2020masked}), which may not perform well under distribution shifts \cite{haresamudram2022assessing}. 

We investigate enhancements to components of the current CPC setup in the field of sensor-based HAR \cite{haresamudram2021contrastive}.
Specifically, we study the encoder architecture, the autoregressive network, and the future prediction tasks of the original setup in order to derive suitable alternatives that can improve HAR performance. 
As in \cite{haresamudram2022assessing}, we perform pre-training on the large-scale Capture-24 dataset, which contains accelerometer data collected at the wrist. 
The evaluation is performed on \emph{six} diverse datasets, where two datasets each contain sensor data from the wrist, waist, and the leg, so as to enable the study of transferring weights across sensor locations. 
Through our experiments, we observe the superior performance of our enhanced CPC over the previous version of CPC, demonstrating the effectiveness of our approach for practical HAR applications.

\label{sec:related}
\section{Related Work}
As detailed in \cite{reiss2012introducing}, conventional, machine-learning based human activity recognition from wearables typically resembles a five-step process, which includes:
\textit{i)} data recording;
\textit{ii)} data pre-processing;
\textit{iii)} segmentation into analysis windows;
\textit{iv)} feature extraction; and
\textit{v)} classification.
In this work, we focus on the fourth stage, and learn representations from large quantities of unlabeled data.
Traditional approaches for unsupervised representation learning include Restricted Boltzmann Machines (RBMs) \cite{plotz2011feature} and varieties of Autoencoders, such as Vanilla, Convolutional, or Recurrent Autoencoders \cite{haresamudram2019role, varamin2018deep}.

In recent years, unsupervised learning has seen a surge in interest, with the introduction of techniques that rely on the design of suitable `pretext' tasks, that aim at capturing relevant aspects of input data.
Such design requires domain expertise and results in representations that are useful for downstream applications. 
This modeling paradigm of `pretrain-then-finetune' has resulted in significant performance boosts for human activity recognition from wearables \cite{saeed2019multi, saeed2021sense, haresamudram2022assessing}.

Multi-task self supervision \cite{saeed2019multi} first introduced self-supervised learning to human activity recognition from sensor data. 
The pretext task involves predicting whether data transformations have been applied or not, in a multi-task setting, thereby modeling the core signal characteristics and sensor behavior under diverse conditions. 
Reconstructing randomly masked out portions of the input was studied for representation learning in \cite{haresamudram2020masked}, where a Transformer \cite{vaswani2017attention} encoder was utilized.
Such a setup aims at capturing the local temporal dependencies, as available from the surrounding context. 

The long term signal or the slowly varying features of the data are captured by predicting not just the next step, but farther into the future in Contrastive Predictive Coding \cite{haresamudram2021contrastive}.
It combines future timestep prediction with contrastive learning, where a longer prediction horizon enables more effective representation learning. 
This setup leverages time-series properties such as the ability to predict future data from the past, and that samples closer in time are more related than those farther apart. 
Therefore, it can be applied to any time-series data in general, and does not require specialized transformations for contrastive learning. 

Another contrastive learning method includes SimCLR \cite{tang2020exploring, chen2020simple, haresamudram2022assessing}, which heavily relies on data transformations \cite{um2017data} as it contrasts two randomly augmented views of the data in a siamese setup, resulting in highly effective representations.
SimSiam \cite{chen2021exploring, haresamudram2022assessing} and Bootstrap Your Own Latent (BYOL) \cite{grill2020bootstrap, haresamudram2022assessing} are also siamese methods, albeit do not contrast the two views.
For multi-sensor setups, ColloSSL \cite{jain2022collossl} and COCOA \cite{deldari2022cocoa} contrast across different sensor locations for training.
Many of the previous methods such as Multi-task self-supervision and the siamese techiques including SimCLR, SimSiam, and BYOL, rely on the design of sutiable data transformations for effective representations. 
Therefore, we focus on CPC, which specifically leverages properties of time-series data for learning features, and thus presents a general framework for time-series data in general.

\label{sec:methodology}
\section{Methodology}
The CPC framework comprises of three components: \emph{(i)} the Encoder; \emph{(ii)} the Aggregator (also referred to as the Autoregressive network \cite{haresamudram2021contrastive, oord2018representation}); and \emph{(iii)} Future timestep prediction.
It learns effective representations by applying contrastive learning in conjunction with future timestep prediction, where predicting farther into the future captures the long-term features of the signal. 
In what follows, we first detail our modifications to the Encoder and Aggregator architectures and the future prediction task, adopting the notations from \cite{haresamudram2021contrastive}.

\begin{figure*}[!t]
    \centering
    \includegraphics[width=0.9\linewidth]{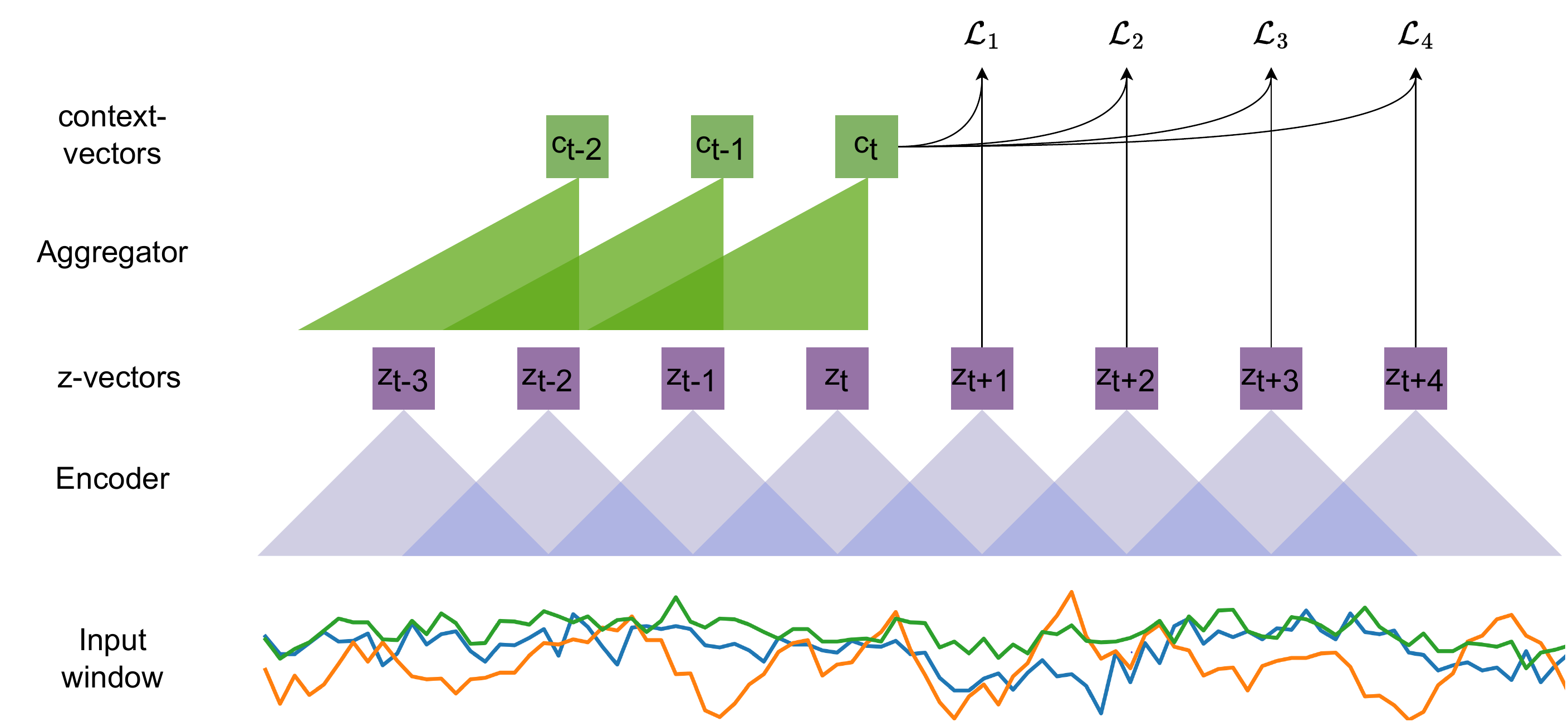}
    \caption{
    Enhancing the Contrastive Predictive Coding (CPC) pipeline: we study whether replacing the encoder with strided convolutions, utilizing a causal convolutional aggregator, and performing future prediction at each timestep results in more effective representation learning for human activity recognition.
    }
    \label{fig:acc2vec}
\end{figure*}

\subsection{Encoder Network}
The Encoder network ($g_{enc}$) utilizes windows of sensor data as input and maps them to latent representations, called $z$-vectors.
It contains four 1D convolutional layers with (32, 64, 128, 256) filters, with a kernel size of (4, 1, 1, 1), and stride of (2, 1, 1, 1).
We utilize reflect padding and the input filter size for the first convolutional layer equals the number of channels of the input, i.e., three, as we utilize triaxial accelerometer data.
Each convolutional layer is followed by ReLU activation and dropout with $p=0.2$. 
Therefore, given a window of data $x$, we obtain $z_m = g_{enc}(x_t)$, obtaining one $z$-vector for every second timestep of sensor data. 
This setup is a modification over the original setup of utilizing three convolutional blocks containing (32, 64, 128) filters and a filter size of 3, resulting in one $z$-vector per timestep of input data. 
We find that the downsampling does not have a substantial negative impact on downstream performance, specifically for locomotion-style activities, while improving computational efficiency.

\subsection{Aggregator Network}
The established CPC framework utilizes a Gated Recurrent (GRU) network to summarize all $z_{\leq t}$ into a context vector $c_t=g_{ar}(z_{\leq t})$.
Similar to \cite{schneider2019wav2vec}, we propose to instead utilize multiple causal convolution blocks to perform the autoregressive summarization, due to its simplicity, faster training times, and ease of parallelization that recurrent models do not enjoy.

We perform parameter tuning over (2, 4, 6) blocks to derive the number of causal convolution layers to utilize (and thus to determine the receptive field of the aggregator).
The kernel sizes of layers are (2, 3, 4, 5, 6, 7), and setting the number of causal convolution blocks to $2$ results in two blocks with the causal convolution filter sizes being (2, 3) respectively.
This is extended for situations where the number of blocks is 4 or 6.
Each convolutional block has 256 filters, with group normalization and ReLU activation, along with skip connections between layers, as suggested in \cite{schneider2019wav2vec}. 

\subsection{Future Time-step Prediction}
Similar to the original formulation of CPC \cite{oord2018representation}, the enhanced version of CPC learns effective representations by training to distinguish a $z$-vector ($z_{t+k}$), that is $k$ timesteps in the future, from a collection of distractor samples drawn from a proposal distribution.
The InfoNCE loss \cite{oord2018representation} is summed over different step sizes and backpropagated to update the network parameters.
For implementation, the CPC framework for HAR \cite{haresamudram2021contrastive}, randomly picks a timestep $t \in [0, T-k]$ from which $k$ subsequent future timesteps are predicted. 
For each future timestep $m < k$, the positive sample is $z_{t+m}$, and the negatives comprise the $z$-vectors at the same timestep $t+m$ from all other windows in the batch. 
This process is computationally more efficient as it does the future prediction only once, thereby improving training times. 
In contrast, we examine whether utilizing the context vector obtained at each timestep to perform future timestep prediction with negatives sampled from the entire batch is more effective for activity recognition (as proposed in \cite{oord2018representation, schneider2019wav2vec}).

\label{sec:settings}
\section{Experiment Settings}
With the (potential) framework enhancements detailed in Section \ref{sec:methodology}, we evaluate model performance on diverse target datasets.
In the following, we provide detailed descriptions of the pre-training and evaluation datasets (Section \ref{sec:datasets}), the implementation details for all methods (Section \ref{sec:implementation}), baseline architectures and parameter tuning protocols (Section \ref{sec:architecture}), and the data preprocessing steps (Section \ref{sec:preprocessing}).

\subsection{Datasets}
\label{sec:datasets}
\subsubsection{Pre-training} 
The Capture-24 dataset is large-scale, containing accelerometer data collected using the Axivity AX3 wrist-worn activity tracker \cite{axivity}.
The sampling rate is 100 Hz, and data were collected in free living conditions.
A Vicon Actigraph wearable camera was utilized for post-hoc labeling, resulting in around 2,500 hours of coarsely labeled data and six broad labels \cite{willetts2018statistical}, namely sleep, sit-stand, mixed, walking, vehicle, and bicycling, as well as 200 fine-grained activities. 
The resulting dataset is imbalanced, with approximately 75\% of the overall dataset covering either sleep or sit-stand. 
The free living conditions as well as the large scale of the dataset make it highly suited for self-supervised learning.

\subsubsection{Evaluation datasets}
In order to study the effectiveness of the approach across sensor locations, we evaluate on \emph{six}, diverse target datasets, where two each have data collected at the wrist, waist, and the leg/ ankle.
The datasets include: HHAR \cite{stisen2015smart},  Myogym \cite{koskimaki2017myogym} at the wrist, Mobiact (v2) \cite{chatzaki2016human}, Motionsense \cite{malekzadeh2018protecting} at the waist, and MHEALTH \cite{banos2014mhealthdroid}, PAMAP2 \cite{reiss2012introducing} at the leg/ ankle.

HHAR, Mobiact, and Motionsense contain primarily locomotion-style activities such as walking, sitting, standing, and going up/down the stairs.
In addition to the locomotion activities, MHEALTH also covers exercises such jogging, cycling, etc., whereas PAMAP2 contains daily living activities such as ironing and vacuum cleaning.
In contrast, Myogym contains fine-grained gym activities such as different types of curls, extensions, raises, rows, etc., with the null class covering over 75\% of the data. 
As in \cite{haresamudram2022assessing}, evaluating across different sensor positions and activities allows us to perform a broad evaluation of the proposed method and understand its strengths and areas of opportunity.

\subsection{Implementation Details}
\label{sec:implementation}
All analysis models were implementated in Pytorch, using the hyperparameter spaces and strategy detailed in \cite{haresamudram2022assessing}.
For Table \ref{tab:diff_locs}, we utilize random search to obtain suitable training parameters, as it is more computationally efficient while also being comparable in performance to grid search \cite{bergstra2012random}.
During pre-training, we randomly choose 20 combinations from the available hyperparameters.

For fine-tuning, we sample 50 combinations across both pre-training and classification, and each combination is evaluated through five-fold cross-validation. 
We average the fine-tuning validation F1-scores across the folds to identify the best performing parameter combination. 
The corresponding average test set F1-score across the folds (for the best combination) is reported in Table \ref{tab:diff_locs}, where five randomized classification runs are performed in order to account for the impact of the random seed.

Both the pre-training and fine-tuning are trained for 50 epochs, and we employ early stopping with a patience of five epochs only for pre-training, given the large size of the data.
As in \cite{haresamudram2022assessing}, the learning rate during classification is decayed by 0.8 every 10 epochs.
The Adam optimizer \cite{kingma2014adam} is utilized (unless otherwise specified), with a batch size of 256.

\subsection{Architecture and Parameters}
\label{sec:architecture}
The parameter search settings for the self-supervised and supervised baselines are identical to \cite{haresamudram2022assessing}, albeit we utilize 10\% of the available training data from Capture-24. 
The actual settings and architectures are also detailed below:

Multi-task self-supervision \cite{saeed2019multi}, and SimCLR \cite{tang2020exploring} utilize the same encoder architecture as introduced in \cite{saeed2019multi}.
It contains three 1D convolutional layers, with (32, 64, 128) filters with kernel sizes of (24, 16, 8),  respectively.
Each convolutional layer is succeeded by the ReLU activation function \cite{nair2010rectified} and dropout \cite{srivastava2014dropout} with $p=0.1$. 
We apply max-pooling after the encoder in order to facilitate classification into activities of interest.
We perform parameter tuning over learning rate and L2 regularization $\in$ \{1e-3, 3e-4, 5e-4\} for Multi-task self-supervision.
In the case of SimCLR, the learning rate $\in$ \{1e-2, 1e-3, 5e-3, 1e-4\}, weight decay $\in$ \{0.0, 1e-4, 1e-5\}, and batch size $\in$ \{1024, 2048, 4096\}.
The SGD optimizer is used instead with a momentum of 0.9 along with a cosine learning schedule.
The projection network contains an MLP with three linear layers of (256, 128, 50) units with intermediate ReLU activation. 
The augmentation functions are obtained from \cite{tang2020exploring} though we do not not chain multiple transforms.

CPC utilizes three 1D convolutional blocks along with a GRU for extracting representations.
Each block comprises a 1D conv. network of (32, 64, 128) filters and kernel sizes of (3, 5) respectively, followed by ReLU and dropout with $p=0.2$. 
The autoregressive network, GRU, has 256 units, 2 layers and dropout of $p=0.2$.
Once again, the learning rate is tuned for $\in$ \{1e-3, 1e-4, 5e-4\}, and L2 regularization $\in$ \{0.0, 1e-4, 1e-5\}, kernel size $\in$ \{3, 5\}.

The Autoencoder \cite{haresamudram2022assessing} has three convolutional blocks for the encoder and the mirror image for the decoder.
The encoder contains (32, 64, 128) filters with ReLU + dropout $p=0.2$; the reverse is used during decoding.
The learning rate $\in$ \{1e-3, 1e-4, 5e-4\}, L2 regularization $\in$ \{0.0, 1e-4, 1e-5\}, and kernel size $\in$ \{3, 5, 7, 9, 11\}.

For the enhanced version of CPC, the learning rate is tuned for $\in$ \{1e-3, 1e-4, 5e-4\}, and L2 regularization $\in$ \{0.0, 1e-4, 1e-5\}, convolutional aggregator layers $\in$ \{2, 4, 6\} layers, the future prediction horizon $\in$ \{10, 12\}, and the number of negatives $\in$ \{10, 15\}. 

For the linear classification, we utilize a single linear layer whereas the MLP comprises three linear layers of size (256, 128, num\_classes) with batch normalization \cite{ioffe2015batch}, ReLU activation, and dropout $p=0.2$ applied in-between.
The classifier is tuned over learning rate $\in$ \{1e-4, 5e-4, 1e-5\} and L2 regularization $\in$ \{0.0, 1e-4, 1e-5\} for all methods apart from Multi-task self-sup. (which instead uses learning rate and L2 regularization \{1e-3, 3e-4, 5e-4\}) and enhanced CPC, which utilizes weight decay.
For clarity all tuned parameter combinations are available at: https://github.com/percom-submission/PerCom-2023/blob/main/README.md

\subsection{Data Pre-Processing}
\label{sec:preprocessing}
We utilize the raw accelerometer data from both the source and target datasets, without any filtering or denoising, as neural networks have the capability of learning directly from raw data \cite{lecun2015deep}.
The Capture-24 dataset is downsampled to 50 Hz, in order to reduce the computational load. 
Similarly, all target datasets (if recorded at higher rates) are downsampled to match the source dataset frequency, as matching the frequency is more effective \cite{haresamudram2022assessing}.
The pre-training dataset is split into train and validation sets by randomly sampling users at a 90:10 ratio. 
As the evaluation is performed only on the target datasets, the pre-training does not have a test split. 
The train set is normalized to have zero mean and unit variance, and the resulting means and variances are applied to normalize the validation set as well.
We also randomly sample 10\% of the available windows for pre-training as \cite{haresamudram2022assessing} showed that the performance is comparable to using the entire train dataset, thereby reducing the computational load. 

Identical to \cite{haresamudram2022assessing}, the window size is set to 2 seconds, with an overlap of 50\% only for the target datasets while Capture-24 has no overlap due to its large scale.
Similar to \cite{haresamudram2022assessing}, we setup five-fold validation for each target dataset.
For the first fold, 20\% of the users are chosen randomly to be the test set, whereas the remaining data is once split by user at an 80:20 ratio into the train and validation sets. 
The second fold (and all remaining folds) are set up such that the test set always contains randomly chosen users who are not part of the test set for any other fold, i.e., across all folds, each user appears in the test set exactly once.
The means and variances computed from the Capture-24 dataset are also utilized on all splits of the target datasets, as it is beneficial for performance \cite{haresamudram2022assessing}.

\begin{table*}[!t]
	\centering
	\caption{
		Activity recognition performance: we report the mean and standard deviation of the five-fold test F1-score across five randomized runs.
		The best performing techniques across datasets are denoted in \green{green}, whereas the best unsupervised methods are indicated in \textbf{bold}.
 	}
	\begin{tabular}{c|cccccc}
	\toprule
	& \multicolumn{2}{c}{Wrist} & \multicolumn{2}{c}{Waist} & \multicolumn{2}{c}{Leg} \\ 
	%
	\multirow{-2}{*}{Method} & HHAR & Myogym & Mobiact & Motionsense & MHEALTH & PAMAP2 \\ 
	\midrule
	\multicolumn{7}{c}{Supervised baselines} \\ 
	\midrule 
	Conv. classifier & 55.63 $\pm$ 2.05 & 38.21 $\pm$ 0.62 & 78.99 $\pm$ 0.38 & 89.01 $\pm$ 0.89 & 48.71 $\pm$ 2.11 & 59.43 $\pm$ 1.56 \\ 
	DeepConvLSTM & 52.37 $\pm$ 2.69 & 39.36 $\pm$ 1.56 & \textbf{\green{82.36 $\pm$ 0.42}} & 84.44 $\pm$ 0.44 & 44.43 $\pm$ 0.95 & 48.53 $\pm$ 0.98 \\ 
	GRU classifier & 45.23 $\pm$ 1.52 & 36.38 $\pm$ 0.60 & 75.74 $\pm$ 0.60 & 87.42 $\pm$ 0.52 & 44.78 $\pm$ 0.47 & 54.35 $\pm$ 1.64 \\ 
	\midrule
	%
	\multicolumn{7}{c}{Self-supervision + Linear evaluation} \\ 
	\midrule 
	Multi-task self.\ sup & 53.71 $\pm$ 1.16 & 24.06 $\pm$ 0.33 & 59.11 $\pm$ 0.41 & 84.80 $\pm$ 0.23 & 36.52 $\pm$ 1.85 & 44.98 $\pm$ 2.29 \\ 
	Autoencoder & 53.93 $\pm$ 0.79 & 14.31 $\pm$ 0.19 & 61.91 $\pm$ 0.22 & 73.34 $\pm$ 0.34 & 26.69 $\pm$ 0.53 & 52.22 $\pm$ 0.96 \\ 
	SimCLR & 52.84 $\pm$ 2.76 & 37.20 $\pm$ 0.36 & 67.15 $\pm$ 0.45 & 87.49 $\pm$ 0.16 & 43.73 $\pm$ 0.82 & 53.74 $\pm$ 0.71 \\ 
	CPC & 59.17 $\pm$ 0.95 & 31.55 $\pm$ 0.18 & 67.52 $\pm$ 0.10 & 83.03 $\pm$ 0.10 & 33.37 $\pm$ 0.46 & 52.39 $\pm$ 0.30 \\ 
	Enhanced CPC & 56.46 $\pm$ 0.93 & 29.66 $\pm$ 0.62 & 71.33 $\pm$ 0.27 & 88.15 $\pm$ 0.32 & 44.69 $\pm$ 0.42 & 56.18 $\pm$ 0.43 \\ 
	\midrule
	%
	%
	\multicolumn{7}{c}{Self-supervision + MLP classifier} \\ 
	\midrule 
	Multi-task self.\ sup & 57.55 $\pm$ 0.75 & 42.73 $\pm$ 0.49 & 72.17 $\pm$ 0.38 & 86.15 $\pm$ 0.42 & 50.39 $\pm$ 0.72 & \textbf{\green{60.25 $\pm$ 0.72}} \\ 
	Autoencoder & 53.64 $\pm$ 1.04 & 46.91 $\pm$ 1.07 & 72.19 $\pm$ 0.35 & 83.10 $\pm$ 0.60 & 40.33 $\pm$ 0.37 & 59.69 $\pm$ 0.72 \\ 
	SimCLR & 56.34 $\pm$ 1.28 & \textbf{\green{47.82 $\pm$ 1.03}} & 75.78 $\pm$ 0.37 & 87.93 $\pm$ 0.61 & 42.11 $\pm$ 0.28 & 58.38 $\pm$ 0.44 \\ 
	CPC & 55.59 $\pm$ 1.40 & 41.03 $\pm$ 0.52 & 73.44 $\pm$ 0.36 & 84.08 $\pm$ 0.59 & 41.03 $\pm$ 0.52 & 55.22 $\pm$ 0.92 \\ 
	Enhanced CPC & \textbf{\green{59.25 $\pm$ 1.31}} & 40.87 $\pm$ 0.50 & \textbf{78.07 $\pm$ 0.27} & \textbf{\green{89.35 $\pm$ 0.32}} & \textbf{\green{53.79 $\pm$ 0.83}} & 58.19 $\pm$ 1.22 \\ 
	\bottomrule
\end{tabular}

	\label{tab:diff_locs}
\end{table*}

\label{sec:results}
\section{Results}
The aim of this work is to derive improvements to the current CPC framework, thereby resulting in improved representation learning from unlabeled wearable sensor data.
As such, we first evaluate the performance of the enhanced CPC framework for activity recognition using a simple linear classifier or a multi-layer perceptron (MLP).
Subsequently, we consider a scenario of paramount importance to wearables, which is when very limited amounts of labeled data are available for fine-tuning. 
Finally, we perform a systematic study of the effect of replacing specific components of the established CPC framework with alternatives and examine any potential improvements. 

\begin{figure*}[t]
    \centering
    \includegraphics[width=0.8\linewidth]{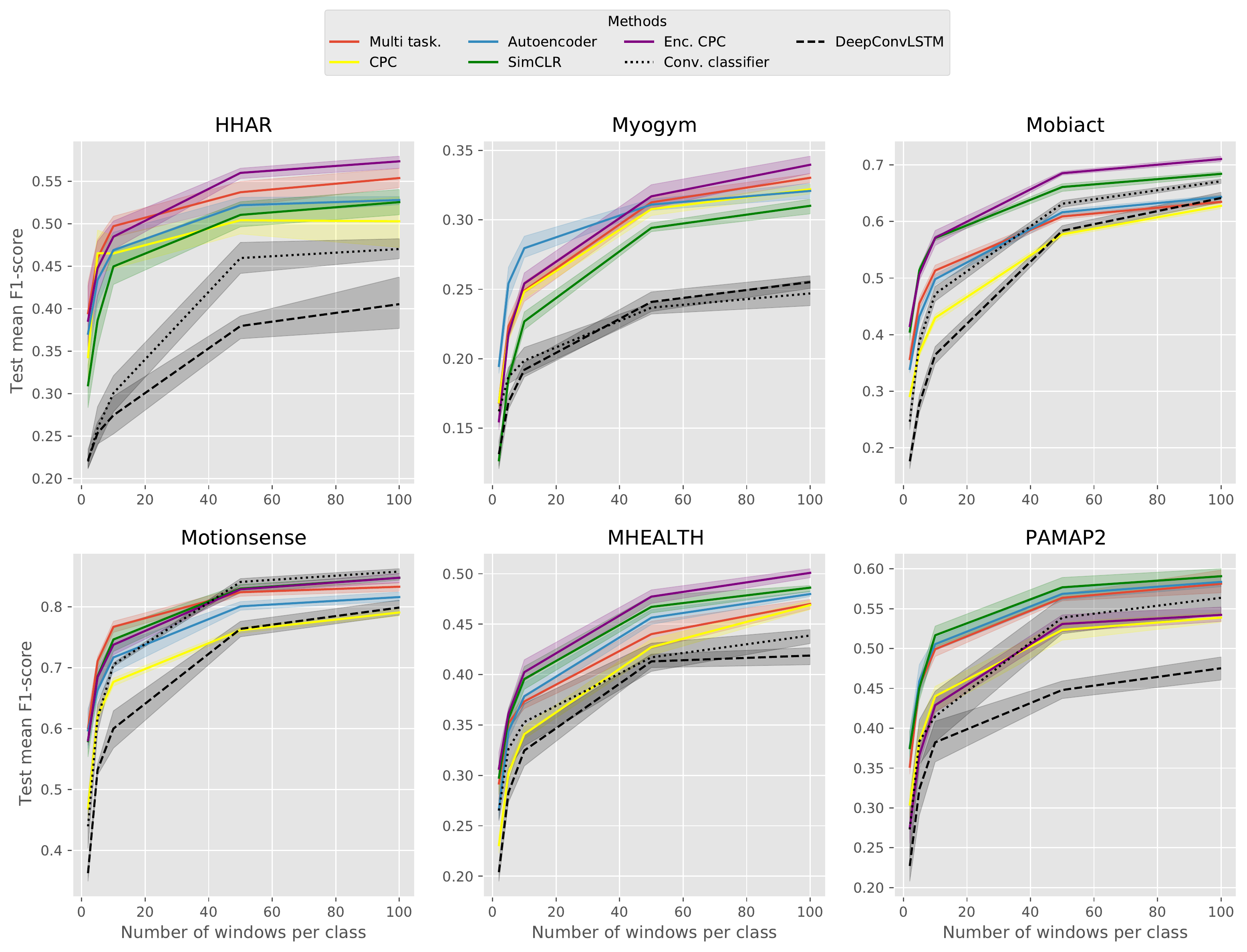}
    \caption{
    Performance for limited annotated data: we study the performance of the enhanced CPC (purple in the figure) when fine-tuning on very few labeled samples per class.
    It is highly effective compared to both supervised and self-supervised learning baselines.
    }
    \label{fig:limited_labels}
\end{figure*}

\subsection{Activity Recognition Performance}
\label{sec:activity_recognition}
Here, we evaluate the effectiveness of enchanced CPC as a feature extractor in the activity recognition chain (ARC) \cite{reiss2012introducing}. 
The performance of the proposed method is contrasted against state-of-the-art self-supervised learning, as well as supervised baselines, which are trained end-to-end, thereby containing considerably more trainable parameters than self-supervised methods, which only optimize the classifier network. 

Following \cite{haresamudram2022assessing},  which contained an assessment of the state-of-the-field regarding self-supervision from wearables, we compare against a diverse set of effective pretext tasks and target datasets:
\emph{(i)} Multi-task self-supervision, which utilizes a 1D convolutional encoder identical to \cite{saeed2019multi}, and learns representations by predicting whether data transformations have been applied or not; 
\emph{(ii)} an Autoencoder, comprising the same encoder architecture as \cite{saeed2019multi}, and training by reconstructing the input after passing through a decoder (which is a mirror image of the encoder);
\emph{(iii)} SimCLR \cite{tang2020exploring,chen2020simple}, whose encoder is the same as \cite{saeed2019multi}, and learns effective weights by contrasting augmented views of the input windows; and,
\emph{(iv)} CPC \cite{haresamudram2021contrastive}, which performs contrastive learning of the future timesteps against distractors sampled from the dataset.

The supervised baselines comprise DeepConvLSTM \cite{ordonez2016deep}, an established network for recognizing activities, a Conv. classifier with the same encoder architecture as \cite{saeed2019multi}, and a simple GRU classifier (Sec.\ \ref{sec:architecture}).

First, we study the performance of the proposed technique across different target dataset sensor locations, results of which are listed in Table \ref{tab:diff_locs}. 
For HHAR, which was collected at the wrist, enchanced CPC + MLP classifier obtains similar performance to CPC + linear classification, while both techniques outperform the supervised baselines by substantial margins. 
On Myogym, both CPC and enhanced CPC perform considerably worse than the SimCLR framework, which shows great improvements over end-to-end training.
The low performance of enhanced CPC on both linear as well as MLP classification can likely be reasoned by the encoder architecture, where the increased striding results in a loss of temporal resolution, thereby negatively impacting the recognition on Myogym (fine-grained gym exercises). 

In the case of the waist-based datasets, we see moderate to good improvements over the current state-of-the-art self-supervised methods.
For Mobiact, we see an increase of around 2\% relative to SimCLR, bringing the performance closer to the Conv.\ classifier as well as DeepConvLSTM. 
We observe a performance boost for Motionsense as well, by around 1.5\%, resulting in similar performance to the best supervised model (Conv).
The stronger performance on the waist-based datasets as well indicates the superior performance of enhanced CPC, as the other methods are not as capable of transferring from wrist$\rightarrow$waist (also observed in \cite{haresamudram2022assessing}). 

Finally, we evaluate the effectiveness of the proposed technique when transferred to leg-based target scenarios.
On MHEALTH, enhanced CPC outperforms the next best self-supervised method (i.e., Multi-task self-sup.) by $>$3\%, showing a significant increase of $>$5\% relative to the the supervised baselines.
In the case of PAMAP2, however, Multi-task self-supervision is more effective for predicting daily living activities, outperforming enhanced CPC by around 2\%. 
It is also able to improve performance relative to the best supervised network, i.e., the Conv. classifier, by around 0.8\%. 
Therefore, we can see that self-supervised learning in general can be more effective than end-to-end training for leg-based applications, and enhanced CPC in particular performs well when detecting exercises at the leg. 

It is also interesting to note that for linear evaluation on wrist-based datasets, enhanced CPC shows a reduction in performance compared to the previous version, yet improves performance for the other sensor locations. 
For Mobiact, Motionsense, and PAMAP2, the increase is around 4\% whereas for MHEALTH, it is over 10\%.
The utilization of the MLP classifier uniformly boosts the performance of enhanced CPC relative to the previous version, by around 4\% on most datasets, and over 10\% on MHEALTH.
Overall, the enhancements introduced for CPC renders it the most effective self-supervised method on four of the six benchmark datasets, indicating its superior performance relative to the baselines, and its general applicability across sensor positions and target activities. 

\begin{table*}[!htbp]
	\centering
	\caption{
	Replacing components of the CPC framework with alternatives and studying impact on performance.	
 	}
	\begin{tabular}{c|cccccc}
	\toprule
	& \multicolumn{2}{c}{Wrist} & \multicolumn{2}{c}{Waist} & \multicolumn{2}{c}{Leg} \\ 
	%
	\multirow{-2}{*}{Setting} & HHAR & Myogym & Mobiact & Motionsense & MHEALTH & PAMAP2 \\ 
	\midrule
	%
	%
	CPC & 55.59 $\pm$ 1.40 & 41.03 $\pm$ 0.52 & 73.44 $\pm$ 0.36 & 84.08 $\pm$ 0.59 & 41.03 $\pm$ 0.52 & 55.22 $\pm$ 0.92 \\ 
	\midrule
	CPC + new encoder & 51.21 $\pm$ 0.93 & 39.68 $\pm$ 0.78 & 71.43 $\pm$ 0.54 & 84.42 $\pm$ 0.45 & 42.06 $\pm$ 0.92 & 53.53 $\pm$ 0.96 \\ 
	CPC + Conv. Agg & 59.42 $\pm$ 0.39 & 42.18 $\pm$ 0.37 & 74.44 $\pm$ 0.35 & 85.92 $\pm$ 0.32 & 50.59 $\pm$ 1.42 & 61.06 $\pm$ 0.72 \\ 
	CPC + new encoder + Conv. Agg & 56.37 $\pm$ 1.51 & 41.39 $\pm$ 0.46 & 75.26 $\pm$ 0.32 & 87.26 $\pm$ 0.58 & 48.58 $\pm$ 1.13 & 59.69 $\pm$ 0.90 \\ 
	\midrule
	Enhanced CPC & 59.25 $\pm$ 1.31 & 40.87 $\pm$ 0.50 & 78.07 $\pm$ 0.27 & 89.35 $\pm$ 0.32 & 53.79 $\pm$ 0.83 & 58.19 $\pm$ 1.22 \\ 
	\bottomrule
\end{tabular}

	\label{tab:ablations}
\end{table*}

\subsection{Performance on Limited Amounts of Labeled Data}
This experiment studies practical scenarios where obtaining large quantities of annotated sensor data may be challenging (or even impossible) due to monetary and/ or privacy concerns. 
They can commonly occur in ubiquitous computing, as real-world deployments may only allow for the collection of a small amount of labeled data.
For example, during device setup, users may be prompted to do a few seconds of orchestrated activities in order to better adapt the prediction model to user-specific patterns and movements.
In such scenarios, training end-to-end only on the limited amount of available data is likely to result in poor performance, as the (typically) complex networks are prone to overfit to the small dataset.
Self-supervised learning is generally more successful in these cases, as the learned weights can function as effective feature extractors, where only the classifier network is optimized with the available data. 

As in Sec. \ref{sec:activity_recognition}, the self-supervised methods are pre-trained using the Capture-24 dataset and only the MLP classifier is updated by backpropagation using the available annotations.
During fine-tuning, we randomly sample \{2, 5, 10, 50, 100\} labeled windows per class from the training set for each fold of the dataset, whereas the validation and test sets are left untouched. 
Once again, we compute the mean of the test set F1-score across the folds and plot the performance across five randomized runs (in order to account for the choice of the randomly chosen windows) in Fig. \ref{fig:limited_labels}. 
We only plot the two best supervised baselines for clarity (as per Table \ref{tab:diff_locs}). 

We observe that for wrist-based datasets (incl. HHAR and Myogym), end-to-end training performs worse than self-supervision, especially when there are $\leq 50$ labeled windows per class available.
For HHAR, enhanced CPC is generally the best performing self-supervised method, especially if at least 25 annotated windows are available.
Interestingly, the simple Autoencoder is more effective when $\leq 25$ labeled windows are available for Myogym; beyond that, the enhancements to CPC are more successful.
In the case of the waist-based datasets, the performance of the supervised baselines is more comparable to many self-supervised methods.
The Conv. classifier is generally quite effective for Mobiact, especially with the availability of increasing quantities of data.
Only SimCLR and enhanced CPC perform better on the dataset, which contains locomotion-style activities along with transition classes. 
For Motionsense however, the Conv. classifier is the better option when we have access to atleast 50 labeled windows.

The effectiveness of the enhancements to CPC (and self-supervision in general) is clearly seen for the leg-based MHEALTH dataset, where it is the most successful technique. 
As in Tab. \ref{tab:diff_locs}, self-supervised approaches such as Multi-task self-sup., Autoencoder, and SimCLR are consistently more effective than fully supervised methods for PAMAP2, with both CPC methods performing worse, and the Conv. classifier obtaining similar performance. 

On the whole, self-supervised methods are very useful in scenarios where limited annotations are available, and enhanced CPC generally performs better than other self-supervised methods. 
This is especially noticeable for the waist-based Mobiact, where it has the capability to significantly improve over the Conv. classifier's performance, even though many other methods do not. 
The consistently high performance of enhanced CPC, across most of the target datasets indicates its applicability in the limited annotated data scenario, thus showcasing its suitability for wearables applications. 
The modifications to the current CPC framework also result in substantial improvements on all target datasets but PAMAP2, where the performance is similar.

\subsection{Investigating the Components of the CPC Framework}
We study three components of the framework: \emph{(i)} the encoder; \emph{(ii)} the autoregressive network; and, \emph{(iii)} the future prediction task. 
In what follows, we examine each component and contrast the performance of the original framework against (potentially more suitable) alternatives. 

\subsubsection{The Encoder}
\label{sec:encoder}
Here, we replace the encoder of the original CPC \cite{haresamudram2021contrastive} framework for wearables (which contains three convolutional blocks) with the encoder of enhanced CPC (that has four convolutional layers with a larger stride) in order to study the effect of the change in architecture.
The striding causes a reduction in the temporal resolution, resulting in a latent representation for every second timestep of data (see Section \ref{sec:methodology}). 
Therefore, we reduce the prediction horizon during pre-training to cover (16, 32) future timesteps and apply the same parameter tuning protocol as described in Section \ref{sec:architecture}, and compare the resulting performance to CPC. 

As seen in Table \ref{tab:ablations} in the row with `CPC + new encoder', replacing the encoder from CPC to the network from enhanced CPC results in a general reduction in performance across all target datasets except MHEALTH, where the performance slightly improves.
For HHAR, we observe the highest drop in performance of 4\% whereas the other datasets see a reduction of 1-2\%.
This shows that the loss of temporal resolution due to the increased striding negatively impacts the downstream recognition performance. 
However, this does not affect Motionsense as the performance remains consistent.
The performance of CPC + new encoder is also significantly lower than enhanced CPC, indicating that the performance improvements are likely not due to the encoder's contributions.
Rather, the reduction in temporal resolution results in quicker training, as the prediction horizon has to be adjusted accordingly.

\subsubsection{The Aggregator/ Autoregressive Network}
\label{sec:aggregator}
Next, we replace the autoregressive network from CPC, the GRU, with the causal convolutions from enhanced CPC.
The primary advantage of this setup is the increased parallelizability due the presence of solely convolutional architecture (across both encoder and the aggregator).
The GRU with 256 units and 2 layers is replaced with causal convolutions tuned over (2, 4, 6) blocks (as in the case of enhanced CPC), in order to obtain the context vector.
The parameter tuning protocol is otherwise identical to Section \ref{sec:architecture} and five random runs of the best performing models are detailed as `CPC + Conv. Agg' in Table \ref{tab:ablations}.

Utilizing the convolutional aggregator in lieu of the GRU results in substantial performance improvements for HHAR and the leg-based MHEALTH and PAMAP2 datasets, however Myogym, Mobiact, and Motionsense show more modest increases. 
In the case of MHEALTH, the improvement is around 9\% whereas PAMAP2 observes an increase of around 6\%. 
The overall performance improvements indicate the superior capability of the causal convolutional aggregator for summarizing the $z$-vectors, and the increased parallelizability makes it more attractive than recurrent networks such as GRUs.

\subsubsection{Future Prediction Task}
In order to study the impact of the future prediction task setup, we first add \emph{both} the new encoder architecture and the causal convolutional aggregator network to the existing CPC framework in the fourth row of Table \ref{tab:ablations} (see `CPC + new encoder + Conv. Agg'). 
The difference between the fourth and fifth rows of Table \ref{tab:ablations} is only the future timestep prediction task, in order to disambiguate its impact on performance.

The addition of the new encoder to `CPC + Conv. Agg' causes a reduction in performance for the wrist- and leg-based datasets; but a modest improvement for the waist-based datasets.
Comparing `Enhanced CPC' to `CPC + new encoder + Conv. Agg', we observe the improvements in performance on four of the benchmark datasets (HHAR, Mobiact, Motionsense, MHEALTH), afforded by the new future prediction task.
Clearly, predicting multiple future timesteps for each context vector across the window, rather than once at a random starting point (as in the CPC framework) helps compensate for the loss in temporal resolution.
While the original encoder can also be utilized with `Enhanced CPC' instead of the new encoder (with increased striding), we observe that the number of future timesteps predicted needs to be increased correspondingly in order to maintain the same prediction horizon (in duration). 
This significantly increases the computational costs, resulting in much higher training times.
Due to this, we utilize the new encoder, as it helps mitigate the increased computational costs due to the new future timestep prediction task.

\label{sec:discussion}
\section{Discussion}
In this paper, we perform an investigation of the components of the CPC framework for human activity recognition from wearables, and aim to derive more effective alternates.
From Table \ref{tab:diff_locs} and Figure \ref{fig:limited_labels}, we observe the overall superior performance of the enhanced CPC for learning representations. 
For datasets with primarily locomotion-style activities, such as HHAR, Mobiact, and Motionsense, the enhanced CPC outperforms the state-of-the-art by substantial margins, indicating its suitability for recognizing activities such as walking, sitting, running, standing, etc., which do not require high temporal resolution. 
In scenarios where fine-grained activities need to be recognized, both CPC and enhanced CPC perform poorly compared to techniques such as SimCLR and Autoencoder. 

The improved performance for waist-based datasets such as Mobiact and Motionsense is also encouraging, as utilizing self-supervised models from the wrist$\rightarrow$waist is typically not as effective as wrist$\rightarrow$wrist (where the location is common) or wrist$\rightarrow$leg (where activities can have tandem movements between the hands and the legs) \cite{haresamudram2022assessing}.
For Motionsense, this results in performance slightly exceeding the supervised baselines whereas Mobiact gets closer in performance to end-to-end training.
The increase however comes at growing computational costs, as predicting future timesteps at each sample in the window is expensive. 
This is somewhat mitigated by the strided convolutional encoder, which reduces the temporal resolution of the encoded data by half.
In situations where training costs are a factor, methods such as SimCLR can be utilized, although it requires knowledge of specialized data transformations.
Further, the original convolutional encoder can also be utilized, although with much longer training times. 
As in \cite{schneider2019wav2vec}, utilizing the causal convolutional aggregator results in a fully convolutional model, increasing the parallelizability of the pre-training, resulting in faster training times.
Finally, we also note that the performance of the enhanced CPC can further be improved by utilizing more data for pre-training and increasing the depth of the encoder and aggregator networks.

\label{sec:conclusion}
\section{Conclusion}
Representation learning, i.e., the fourth stage of the activity recognition chain, is of paramount importance for automatically recognizing activities from wearable sensor data.
As such, the availability of large-scale datasets of human movements has facilitated the design and study of methods that can leverage unlabeled data to perform effective representation learning. 
In this paper, we investigated improvements to Contrastive Prediction Coding (CPC) framework for wearables-based human activity recognition.
We proposed suitable replacements to the encoder network, the aggregator network, and the future timestep prediction task, so as to improve downstream activity recognition performance.
We modified the encoder to have higher striding, replaced the GRU aggregator with causal convolutions in order to summarize the latent vectors into a context vector, and performed future timestep prediction for each context vector.
These changes result in performance substantial improvements on four of six benchmark datasets, even when transferring across target sensor locations.
Especially when there is paucity of labeled data, the enhanced version of CPC (and self-supervision in general) significantly outperforms supervised learning baselines, demonstrating the promise of the `pretrain-then-finetune' paradigm for representation learning on wearables applications.


\bibliographystyle{IEEEtran}
\bibliography{refs}

\end{document}